\documentclass[final]{cvpr}

\usepackage{times}
\usepackage{epsfig}
\usepackage{graphicx}
\usepackage{amsmath}
\usepackage{amssymb}
\usepackage{float} % @dv
\usepackage{adjustbox} % @sh
\newcommand{\bftab}{\fontseries{b}\selectfont} % @sh
\usepackage{pifont} % @sh: for \xmark
\newcommand{\xmark}{\ding{55}} % @sh
\usepackage{graphicx}
\usepackage{multirow}

\usepackage[pagebackref=true,breaklinks=true,colorlinks,bookmarks=false]{hyperref}

\def\cvprPaperID{10600} % *** Enter the CVPR Paper ID here
\def\confYear{CVPR 2021}
\begin{document}

%%%%%%%%% TITLE
\title{FloodNet: A High Resolution Aerial Imagery Dataset for Post Flood Scene Understanding}

\author{Maryam Rahnemoonfar, Tashnim Chowdhury, Argho Sarkar, Debvrat Varshney, Masoud Yari\\
University of Maryland Baltimore County\\
Baltimore, Maryland\\
{\tt\small maryam@umbc.edu, tchowdhury@umbc.edu, asarkar2@umbc.edu, dvarshney@umbc.edu, yari@umbc.edu}

\and
Robin Murphy\\
Texas A\&M University\\
College Station, Texas\\
{\tt\small murphy@cse.tamu.edu}

}

\iffalse
\newcommand\myfigure{%
\centering
    %\rule{2cm}{2cm}
    \begin{center}
		\includegraphics[width=\linewidth]{./image/harvey-annotation-example.png}
	\end{center}
\captionof{figure}{Some stuff about the teaser}
\label{fig:teaser}
}
\fi
% \cvprPaperID{}

\maketitle
%%%%%%%%% ABSTRACT

\begin{figure*}
\centering
        \includegraphics[scale=.6]{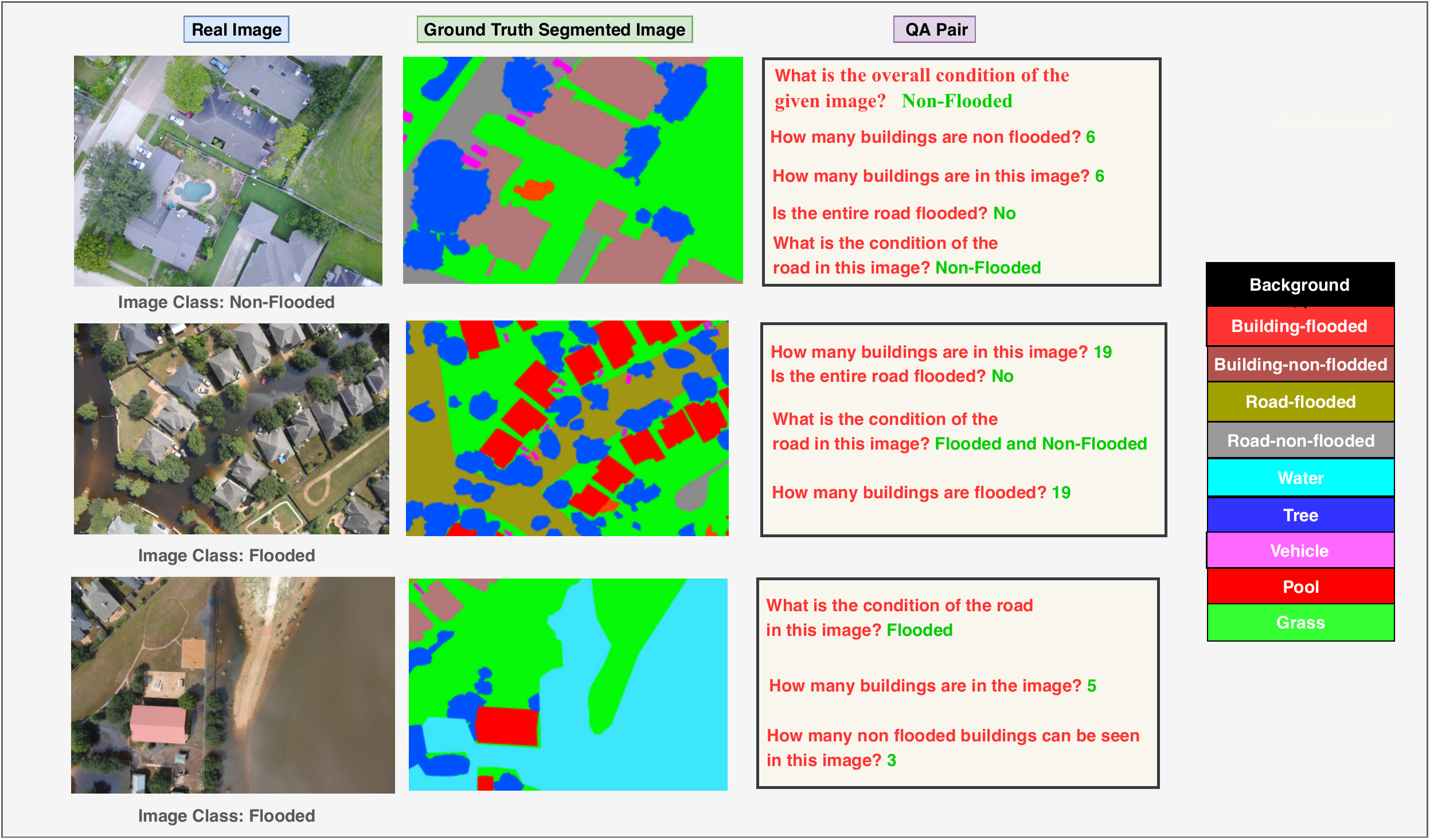}
    % \label{Fig:data_rep_vqa}
   
    \caption{FloodNet dataset overview for Classification, Semantic Segmentation and Visual Question Answering }
    \label{fig:data_rep}
\end{figure*}

\begin{abstract}
   Visual scene understanding is the core task in making any crucial decision in any computer vision system. Although popular computer vision datasets like Cityscapes, MS-COCO, PASCAL provide good benchmarks for several tasks (e.g. image classification, segmentation, object detection), these datasets are hardly suitable for post disaster damage assessments. On the other hand, existing natural disaster datasets include mainly  satellite  imagery  which  have low  spatial resolution and a high revisit period. Therefore, they do not have a scope to provide quick and efficient damage assessment tasks. Unmanned  Aerial  Vehicle (UAV) can effortlessly access difficult places during any disaster and collect high resolution imagery  that is required for aforementioned tasks of computer vision. To address these issues we present a high resolution UAV imagery, FloodNet, captured  after the hurricane  Harvey. This dataset demonstrates the post flooded damages of the affected areas. The images are labeled pixel-wise for semantic segmentation task and questions are produced for the task of visual question answering. FloodNet poses several challenges including detection of flooded roads and buildings and distinguishing between natural water and flooded water. With the advancement of deep learning algorithms, we can analyze the impact of any disaster which can make a precise understanding of the affected areas. In this paper, we compare and contrast the performances of baseline methods for image classification, semantic segmentation, and visual question answering on our dataset.
\end{abstract}

% \begin{figure*}[!ht]
% 	\begin{center}
% 		\includegraphics[width=\linewidth]{./image/harvey-annotation-example.PNG}
% 	\end{center}
% 	\caption{Illustration of of FloodNet dataset. First row shows original image and the second row shows the corresponding annotations.}
% 	\label{fig:floodnet-annotation-exm}
% \end{figure*}

\section{Introduction}
%\input{intro_argho}
% As a result of a significant increase in the earth's surface temperature, 
% numerous calamities have become a more frequent phenomenon, such as hurricanes, destructive floods, wildfires. 
% By damaging our environment, these disasters produce a lot of detrimental effects on our lives. Damages and debris are the consequences of natural disasters. We can provide assistance that can save many lives and rebuild the affected areas by understanding and identifying the extent of damage in affected areas. Traditionally manual human activities that are laborious, as well as expensive, are rendered for data collection purposes during emergencies. However, Unmanned Aerial Vehicles (UAVs) can easily reach difficult locations and capture high-resolution images which show the real-time condition of the affected environments. 

% As a consequence, we can incorporate multiple computer vision tasks such as classification, segmentation, visual question answering to explain the situation very precisely. With this regard, FloodNet dataset provides classification, semantic segmentation, and visual question answering tasks for post disaster damage assessment. 

% Figure \ref{Fig:data_rep_vqa} represents our dataset. 

%\begin{figure*}[!htp]
%\centering
%        \includegraphics[scale=.6]{image/data_rep.pdf}
%    \label{Fig:data_rep_vqa}
%   
%    \caption{FloodNet Dataset Overview for Semantic Segmentation and Visual Question Answering}
%\end{figure*}

Understanding of a visual scene from images has the potential to advance many decision support systems. The purpose of scene understanding is to classify the overall category of scene as well as constituting interrelationship among different object classes at both instance and pixel level. Recently, several datasets \cite{5206848, lin2015microsoft, Everingham15} have been presented to study different aspects of scenes by implementing many computer vision tasks. A major factor in success of most of the deep learning algorithms is the availability of large-scale dataset. Publicly available ground imagery datasets such as ImageNet\cite{5206848}, Microsoft COCO\cite{lin2015microsoft}, PASCAL VOC\cite{Everingham15}, Cityscapes\cite{cordts2016cityscapes}  accelerate the advanced development of current deep neural networks, but the annotation of aerial imagery is scarce and more tedious to obtain.

%However, different features of a scene changes dramatically after any natural calamities like flood, hurricane compare to pre-disaster condition. 
Aerial scene understanding dataset are helpful for urban management, city planning, infrastructure maintenance, damage assessment after natural disasters, and high definition (HD) maps for self-driving cars. Existing aerial datasets, however, are limited mainly to classification \cite{gupta2019creating, kyrkou2019deep} or semantic segmentation \cite{gupta2019creating, rudner2019multi3net} of few individual classes such as roads or buildings. Moreover, all of these datasets are collected in normal conditions and computer vision algorithms are mainly developed for normal looking objects.
Most of these datasets do not address the unique challenges in understanding post disaster scenarios as a task  for disaster damage assessment. For quick response and recovery in large scale after a natural disaster such as hurricane, wildfire, and extreme flooding access to aerial images are critically important for the response team. 
To fill this gap we present \textit{FloodNet} dataset associated with three different computer vision tasks namely classification, semantic segmentation, and visual question answering. 
% \par Researchers are making success in applying algorithms 
% for computer vision for the greater good. 
% Therefore a lot efforts  in   agricultural, health, and environment \cite{efremova2019ai,m2019semantic,wu2019deep,schmidt2019visualizing,ronneberger2015u} have already been made. These have 
% shown the prospect of a large-scale computer vision for social good. 
% Computer vision tasks for post-disaster damage assessment, 
% however, were not considered to a large degree. 
\par Although several datasets \cite{bischke2017multimedia,benjamin2018multimedia,gupta2019xbd,weber2020eccv} are provided for post disaster damage assessments, they have numerous issues to tackle.  Most of those datasets contain satellite images and images collected from social media.  Satellite images  are low in resolution and captured from high altitude. They are affected from several noises including clouds and smokes. Moreover, deploying satellites and collecting images from these are costly. On the other hand, images posted on social media are noisy and not scalable for deep learning models. To address this issues, our dataset, \textit{FloodNet}, provides high resolution images taken from low altitude. These characteristics of \textit{FloodNet} brings more clarity to scenes and thus help deep learning models in making more accurate decisions regarding post disaster damage assessment. In addition, most of tasks considering natural disaster datasets are restricted to mainly classification and object detection. Our dataset offers  advanced computer vision challenges namely semantic segmentation and visual question answering besides classification. All these three computer vision tasks  can provide assistance in complete understanding of a scene and help rescue team to manage their operation efficiently during emergencies. Figure \ref{fig:data_rep} shows sample annotations offered by FloodNet.

% \par To make high-level understanding of affected region as well as damages, we  have considered Semantic Segmentation and Visual Question Answering task for our dataset. Semantic segmentation is the task of assigning semantic labels to every pixels of an image which is a core part of image understanding. Visual Question Answering is regarded as a cognitive activity that separates it from other perceptual activities, such as the classification of images. A model needs to detect objects for any VQA task as well as classify their attributes and figure out the interactive relationship. 

\par Our contribution is two folds. First we introduce a high resolution UAV imagery named \textit{FloodNet} for post disaster damage assessment. Secondly, we compare the performance of sevral classification, semantic segmentation and visual question answering on our dataset. To the best of our knowledge, this is the first VQA work focused on UAV imagery for any disaster damage assessment.

The reminder of this paper is organized as follows: it begins
with highlighting the existing datasets for natural disaster, semantic segmentation, and visual question answering in section \ref{sec:related_work}.
Next, section \ref{sec:dataset} describes the \textit{FloodNet} dataset including its collection and annotation process. Section \ref{sec:exp} describes the experimental setups for all three aforementioned tasks along with complete result analysis of corresponding tasks.  Finally section \ref{sec:final} summarizes the results
including conclusion and future works. 
% This gives us a complete understanding of the scene and helps 
% us to make the right decision at the moment of an emergency.  

%\input{table/vqa_dataset}
%-------------------------------------------------------------------------

\section{Related Works}
\label{sec:related_work}

In this section we provide an overview of datasets designed for natural disasters damage analysis, followed by
a survey of techniques targeting aerial and satellite image classification, segmentation, and VQA.

\begin{table*}
\caption{A brief summary of existing datasets.} \label{tab:dataset-summary}
%\vspace{.2cm}
\scalebox{0.65}{

\begin{tabular}{|c|c|c|c|c|c|c|c|}
\hline
\textbf{Dataset}         & \textbf{Types of Images}                    & \textbf{UAV imagery} & \textbf{Post Disaster} & \textbf{Resolution of Images} & \textbf{Classification} & \textbf{Semantic Segmentation} & \textbf{VQA}                                                                            \\ \hline
ImageNet \cite{5206848}                  & Real-world images                           & No                                 & No                                      & average 400 × 350  &  \color{green}\checkmark & \color{red} \xmark & \color{red} \xmark                                                                                                     \\ \hline
Cityscapes \cite{lin2015microsoft}                  & Real-world images                           & No                                 & No                                      & 1280 × 720  & \color{red} \xmark & \color{green} \checkmark & \color{red} \xmark                                                                                                     \\ \hline

DAQUAR \cite{malinowski2014multi}                  & Real-world images                           & No                                 & No                                      & 640 × 480  & \color{red} \xmark & \color{red} \xmark & \color{green} \checkmark                                                                                                     \\ \hline
COCO-QA \cite{ren2015exploring}                 & Real-world images                           & No                                 & No                                      & 640 × 480 & \color{red} \xmark & \color{red} \xmark & \color{green} \checkmark                                                                                   \\ \hline
COCO-VQA \cite{antol2015vqa}                & Real  world images, abstract cartoon images & No                                 & No                                      & 640 × 480   & \color{red} \xmark & \color{red} \xmark & \color{green} \checkmark                                                                                                     \\\hline
Visual Genome \cite{krishna2017visual}          & Real-world images                           & No                                 & No                                      & varies in size & \color{red} \xmark & \color{red} \xmark & \color{green} \checkmark \\ \hline
Visual7W   \cite{zhu2016visual7w}               & Real-world images                           & No                                 & No                                      & varies in size   & \color{red} \xmark & \color{red} \xmark & \color{green} \checkmark                                                                                      \\ \hline
TDIUC   \cite{kafle2017analysis}                 & Real-world images                           & No                                 & No                                      & varies in size  & \color{red} \xmark & \color{red} \xmark & \color{green} \checkmark                                                                              \\ \hline
CLEVR \cite{johnson2017clevr}                    & Geometrical Shape                           & No                                 & No                                      & 320 x 240 (in default settings)    & \color{red} \xmark & \color{red} \xmark & \color{green} \checkmark                                                                              \\ \hline
PATHVQA \cite{he2020pathvqa}                 & Medical Images                              & No                                 & No                                      & Varies in size   & \color{red} \xmark & \color{red} \xmark & \color{green} \checkmark                                                                                                \\ \hline
 VQA-MED \cite{abacha2019vqa}       & Medical Images                              & No                                 & No                                      & Varies in size    & \color{red} \xmark & \color{red} \xmark & \color{green} \checkmark                                                                                               \\ \hline
 
Nguyen et al. \cite{nguyen2017automatic}       & Post Disaster Images                              & No                                 & Yes                                      & Varies in size    & \color{green} \checkmark & \color{red} \xmark & \color{red} \xmark                                                                                               \\ \hline

ABCD \cite{fujita2017damage}    & Pre and Post Disaster Images  & No                                 & Yes                  & Varies in size   & \color{green} \checkmark & \color{red} \xmark & \color{red} \xmark               \\ \hline

SpaceNet + Deepglobe \cite{doshi2018satellite}     & Pre and Post Disaster Images 
& No                                 & Yes                           & Varies in size  & \color{red} \xmark & \color{green} \checkmark & \color{red} \xmark       \\ \hline

Chen et al. \cite{chen2018benchmark}     & Post Disaster Images 
& No                                 & Yes                           & Varies in size & \color{red} \xmark & \color{red} \xmark & \color{red} \xmark      \\ \hline

OSCD \cite{daudt2018urban}     & Urban Change Images 
& No                                 & No                           & Varies in size & \color{red} \xmark & \color{red} \xmark & \color{red} \xmark      \\ \hline

fMoW \cite{christie2018functional}     & Pre and Post Disaster Images 
& No                                 & Yes                           & Varies in size  & \color{green} \checkmark & \color{red} \xmark & \color{red} \xmark     \\ \hline 

AIDER \cite{kyrkou2019deep}     & Post Disaster Images 
& Yes                                 & Yes                           & Varies in size & \color{green} \checkmark & \color{red} \xmark & \color{red} \xmark      \\ \hline

Rudner et al. \cite{rudner2019multi3net}     & Post Disaster Images 
& No                                 & Yes                           & Varies in size & \color{red} \xmark & \color{green} \checkmark & \color{red} \xmark      \\ \hline

xBD \cite{gupta2019creating}    & Pre and Post Disaster Images 
& No                                 & Yes                           & Varies in size & \color{green} \checkmark & \color{green} \checkmark & \color{red} \xmark      \\ \hline

ISBDA \cite{zhu2020msnet}    & Post Disaster Images 
& Yes                                 & Yes                           & Varies in size  & \color{red} \xmark & \color{red} \xmark & \color{red} \xmark     \\ \hline

\textbf{FloodNet (Ours)} & \textbf{Post Disaster Images}    & \textbf{Yes}                       & \textbf{Yes}                            & \textbf{4000× 3000}   & \color{green} \checkmark & \color{green} \checkmark & \color{green} \checkmark                                                                                           \\ \hline
\end{tabular}}
\end{table*}

%%----------------------Tashnim---------------------------
\subsection{Datasets}

Natural disaster dataset can be initially classified into two classes: A) Non-imaging dataset (text, tweets, social media post) \cite{imran2015processing, reuter2018fifteen} and B) Imaging datasets \cite{rudner2019multi3net, gupta2019creating, chen2018benchmark}. Based on the image capture position existing imaging natural disaster datasets can be further classified into three classes: B1) Ground-level images \cite{nguyen2017damage}, B2) Satellite imagery \cite{chen2018benchmark, gupta2019creating, doshi2018satellite, daudt2018urban, christie2018functional, rudner2019multi3net}, and B3) Aerial imagery \cite{kyrkou2019deep, zhu2020msnet, fujita2017damage}. Recently several datasets have been introduced by researchers for natural disaster damage assessment. Nguyen et al. proposed an extension of AIDR system \cite{nguyen2017automatic} to collect data from social media in \cite{nguyen2017damage}. AIST Building Change Detection (ABCD) dataset has been proposed in \cite{fujita2017damage} which includes aerial post tsunami images to identify whether the buildings have been washed away. A combination of SpaceNet \cite{cosmiqworksnvidia} and DeepGlobe \cite{demir2018deepglobe} was presented in \cite{doshi2018satellite} and a segementation model was proposed to detect changes in man-made structures to estimate the impact of natural disasters.  Chen et al. in \cite{chen2018benchmark} proposed a fusion of different data resources for automatic building damage detection after a hurricane. The dataset includes satellite and aerial imageries along with vector data. Onera Satellite Change Detection (OSCD) dataset was proposed in \cite{daudt2018urban} which consists of multispectral aerial images to detect urban growth and changes with time. A collection of images of buildings and lands named Functional Map of the World (fMoW) was introduced by Christie et al. in \cite{christie2018functional}. Aerial Image Database for Emergency Response (AIDER) is proposed by Kyrkou et al. in \cite{kyrkou2019deep} for classification of UAV imagey. Rudner et al. \cite{rudner2019multi3net} propose a satellite imagery collected from Sentinel-1 and Sentinel-2 satellites for semantic segmentation of flooded buildings. Gupta et al. proposed xBD \cite{gupta2019creating} which have both pre- and post-event satellite images in order to assess building damages. Recently ISBDA (Instance Segmentation in Building Damage Assessment) is created by Zhu et al. in \cite{zhu2020msnet} for instance segmentation while images are collected using UAVs.

A comparative study among different disaster and non disaster datasets is shown in Table \ref{tab:dataset-summary}. As you can see in Table \ref{tab:dataset-summary}, our dataset is the only high resulting UAV dataset collected after a hurricane which contains all computer vision tasks including classification, semantic segmentation, and VQA.  Although several pre- and post-disaster datasets have been proposed over the years, these datasets are primary satellite imageries. Satellite imageries, including those with high resolution, do not provide enough details about the post disaster scenes which are necessary to distinguish among different damage categories of different objects. On the other hand the primary source of the ground-level imageries is social media \cite{nguyen2017damage}. These imageries lack geo location tags \cite{zhu2020msnet} and suffers from data scarcity for deep learning training \cite{weber2020eccv}.  Although some aerial datasets  \cite{kyrkou2019deep, zhu2020msnet} are prepared using UAVs, these datasets lack low altitude high resolution images. AIDER \cite{kyrkou2019deep} dataset collected images from different sources for image classification task and contains far more examples of normal cases rather than damaged objects; therefore lacks consistency and generalization. ISBDA \cite{zhu2020msnet} provides only building instance detection capability rather than inclusion of other damaged objects and computer vision tasks like semantic segmentation and VQA. To address all these issues, FloodNet includes low altitude high resolution post disaster images annotated for classification, semantic segmentation, and VQA. FloodNet provides more details about the scenarios which help to estimate the post disaster damage assessment more accurately.

\subsection{Algorithms}
Here we review the related algorithms and some of their applications in disaster damage assessment.
%%--------Debvrat-----------------------
\subsubsection{Classification}
The utility of Deep Neural Networks was realized when they achieved high accuracy in categorizing images into different classes. This was given a boost mainly by AlexNet \cite{krizhevsky2017imagenet} which achieved state-of-the-art performance on the ImageNet \cite{deng2009imagenet} dataset in 2012. As this is arguably the most primitive computer vision task, a lot of networks were proposed subsequently which could perform classification on public datasets such as CIFAR\cite{cifar10,krizhevsky2009learning}, MNIST\cite{lecun2010mnist}, and FashionMNIST \cite{xiao2017fashion}.

This led to a rise in networks such as \cite{simonyan2014very}, \cite{he2016deep}, \cite{szegedy2015going}, \cite{chollet2017xception}, \cite{howard2017mobilenets} etc., where the network architectures were experimented with different skip connections, residual learning, multi-level feature extraction, separable convolutions, and optimization methods for mobile devices. Although these networks achieved good performance on day to day images of animals and vehicles, they were hardly sufficient to make predictions on scientific datasets such as those captured by aerial or space borne sensors. 

In this regard, some image classification networks have been explored for the purpose of post-disaster damage detection \cite{kyrkou2020emergencynet,nguyen2017automatic,Bejiga_2017,sharma2017deep,zhao2018saliency}. \cite{nguyen2017automatic} used crowd sourced images from social media which captured disaster sites from the ground level. \cite{Bejiga_2017} used a Support Vector Machine on top of a Convolutional Neural Network (CNN) followed by a Hidden Markov Model post-processing to detect avalanches. \cite{sharma2017deep} compared \cite{simonyan2014very} and \cite{he2016deep} for fire detection, but then again the dataset used contained images taken by hand-held cameras on the ground. \cite{zhao2018saliency} developed a novel algorithm which focused on wildfire detection through UAV images. \cite{kyrkou2020emergencynet} have done  extensive work by developing a CNN for emergency response towards fire, flood, collapsed buildings, and crashed cars. Our paper can contribute in this domain by providing multi feature flooded scenes that can inspire the efficient training of more neural networks.

%%--------Tashnim-----------------------
\subsubsection{Semantic segmentation}
Semantic segmentation is one of the prime research area in computer vision and an essential part of scene understanding. Fully Convolutional Network (FCN) \cite{long2015fully} is a pioneering work which is followed by several state-of-art models to address semantic segmentation. \iffalse From the perspective of design architecture existing segmentation models can be classified into two classes: encoder-decoder based methods and pyramid pooling based methods. \fi From the perspective of contextual aggregation, segmentation models can be divided into two types. Models, such as PSPNet \cite{zhao2017pyramid} or DeepLab \cite{chen2017deeplab, chen2017rethinking} perform spatial pyramid pooling \cite{grauman2005pyramid, lazebnik2006beyond} at several grid scales and have shown promising results on several segmentation benchmarks. The encoder-decoder networks combines mid-level and high-level features to obtain different scale global context. Some notable works using this architecture are \cite{chen2017rethinking, ronneberger2015u}. On the other hand, there are models \cite{zhao2018psanet, zhang2018context, fu2019dual} which obtain feature representation by learning contextual dependencies over local features.

Besides proposing natural disaster datasets many researchers have also presented different deep learning models for post natural disaster damage assessment. Authors in \cite{doshi2018satellite} perform previously proposed semantic segmentation \cite{doshi2018residual}  on satellite images to detect changes in the structure of various man-made features, and thus detect areas of maximal impact due to natural disaster. Rahnemoonfar et al. present a densely connected recurrent neural network in \cite{rahnemoonfar2018flooded} to perform semantic segmentation on UAV images for flooded area detection. Rudner et al. fuse multiresolution, multisensor, and multitemporal satellite imagery and propose a novel approach named Multi3Net in \cite{rudner2019multi3net} for rapid segmentation of flooded buildings. Gupta et al. propose a DeepLabv3 \cite{chen2017rethinking} and DeepLabv3+ \cite{chen2018encoder} inspired RescueNet in \cite{gupta2020rescuenet} for joint building segmentation and damage classification. All these proposed methods address the semantic segmentation of specific object classes like river, buildings, and roads rather than complete scene post disaster scenes. \iffalse In this paper we use three state-of-art segmentation methods, ENet \cite{paszke2016enet}, DeepLabv3+ \cite{chen2018encoder}, and PSPNet \cite{zhao2017pyramid} to perform semantic segmentation not only on buildings and roads, but also other objects of an image including pools, trees, vehicles, and natural water. \fi

Above mentioned state-of-art semantic segmentation models have been primarily applied on ground based imagery \cite{cordts2016cityscapes, mottaghi2014role}. In contrast we apply three state-of-art semantic segmentation networks on our proposed FloodNet dataset. We adopt one encoder-decoder based network named ENet \cite{paszke2016enet}, one pyramid pooling module based network PSPNet \cite{zhao2017pyramid}, and the last network model DeepLabv3+ \cite{chen2018encoder} employs both encoder-decoder and pyramid pooling based module.

\subsubsection{Visual Question Answering}
%%----------------------Argho-----------------
%\input{related-work_argho}

Many researchers proposed several datasets and methods for Visual Question Answering task. However, there are no such datasets  apt for training and evaluating VQA algorithms regarding disaster damage assessment tasks.
\par  To find the right answer, VQA systems need to model the question and image (visual content). Substantial research efforts have been made on the VQA task based on real natural and medical imagery  in the computer vision and natural language processing communities \cite{antol2015vqa,  yang2016stacked, kim2018bilinear, gao2019dynamic} using deep learning-based multimodal methods \cite{lu2016hierarchical,  xu2016ask,fukui2016multimodal, anderson2018bottom,yu2019multi,yu2018beyond,ben2017mutan,kim2016hadamard, yu2017multi}. In these methods, different approaches for the fined-grained fusion between semantic features of image and question have been proposed. Most of the recent VQA algorithms have trained on natural image based datasets such as DAQUAR\cite{10.1007/978-3-642-33715-4_54}, COCO-VQA \cite{antol2015vqa}, Visual Genome\cite{krishna2017visual}, Visual7W \cite{zhu2016visual7w}. In addition Path-VQA \cite{he2020pathvqa} and VQA-MED \cite{abacha2019vqa} are medical images for which VQA algorithms are also considered. In this work, we present \textit{FloodNet} dataset to build and test  VQA algorithms that can be implemented during natural emergencies. To the best of our knowledge, this is the first VQA dataset focused on UAV imagery for disaster damage assessment. To evaluate the performances of existing VQA algorithms  we have implemented baseline models, Stacked Attention network\cite{yang2016stacked}, and MFB with Co-Attention\cite{yu2017multi} network on our dataset.
\section{The FloodNet Dataset}
\label{sec:dataset}
%%----------------------- Argho---------------

The data is collected with small UAV platform, DJI Mavic Pro quadcopters, after \textit{Hurricane Harvey}. Hurricane Harvey made landfall near Texas and Louisiana on August, 2017, as a Category 4 hurricane. The Harvey dataset consists of video and imagery taken from several flights conducted between August 30 - September 04, 2017, at Ford Bend County in Texas and other directly impacted areas. The dataset is unique for two reasons. One is fidelity: it contains imagery from sUAV taken during the response phase by emergency responders, thus the data reflects what is the state of the practice and can be reasonable expected to be collected during a  disaster. Second: it is the only known database of sUAV imagery for disasters. Note that there are other existing databases of imagery from unmanned and manned aerial assets collected during disasters, such as National Guard Predators or Civil Air Patrol, but those are larger, fixed-wing assets that operate above the 400 feet AGL (Above Ground Level), limitation of sUAV. All flights were flown at 200 feet AGL, as compared to manned assets which normally fly at 500 feet AGL or higher. 
%We have collected images from the flood-affected areas after \textit{Hurricane Harvey} with the assistance of an unmanned aerial vehicle ( UAV). 
Such images are very high in resolution, making them unique compared to other data sets for natural disasters. The post-flooded damages to affected areas are demonstrated in all the images. There are several objects (e.g. construction, road) and related attributes ( e.g. state of an object such as flooded or non-flooded after Hurricane Harvey) represented by these images. For the preparation of this dataset for semantic segmentation and visual question answering, these attributes are considered. 

\subsection{Annotation Tasks}
\label{sec:data-overview-challenge}
After natural disasters, the response team first need to identify the affected neighborhoods such as flooded neighborhoods (classification tasks). Then on each neighborhood they need to identify flooded buildings and roads (semantic segmentation) so the rescue team can be sent to affected areas. Furthermore, damage assessment after any natural calamities done by querying about the changes in object’s condition so they can allocate the right resources. Based on these needs and with the help of response and rescue team, we defined classification, semantic segmentation and VQA tasks.    
%Collected data are annotated on V7 Darwin platform \cite{V7Darwin} for semantic segmentation. 
In total 3200 images have been annotated with 9 classes which include building-flooded, building-non-flooded, road-flooded, road-non-flooded, water, tree, vehicle, pool, and grass. A buildings is classified as flooded when at least one side of a building is touching the flood water. Although we have classes created for flooded buildings and roads, to distinguish between natural water and flood water, ``water'' class has been created which represents any natural water body like river and lake. For the classification task, each image is classified either ``flooded'' or ``non-flooded''. If more than $30 \%$ area of an image is occupied by flood water then that area is classified as flooded, otherwise non-flooded. Number of images and instances corresponding to different classes are shown in Table \ref{table:harvey-data-stat}. 
our images are quite dense. On average, it take about one hour to annotate each image.  To ensure
high quality, we performed the annotation process iteratively with a two-level quality check over each class. The images are annotated on V7 Darwin platform \cite{V7Darwin} for classification and semantic segmentation. We split the dataset into training, validation, and test sets with $70 \%$ for training and $30 \%$ for validation and testing. The training, validation, and testing sets for all the three tasks will be publicly available. 

\iffalse
There are several challenges have been posed by FloodNet for any computer vision tasks. Among these challenges detecting flooded building is the prime one. Since UAV images only include top view of a building, it is very difficult to estimate how much damages are done on that building. Horizontal view brings most information about a building's current condition after the flood, but the UAV images provide only top views. We are defining a building to be flooded if at least one end or any outside wall of the building is touching the flood water. From an UAV image it is very difficult to verify whether flood water is touching at least one end of a building which makes distinguishing between flooded and non-flooded buildings a challenging task for any network model. The next difficult tasks are to detect vehicles and pools. Among all the existing classes, vehicles and pools are the smallest in shape and therefore would be difficult for any network model to detect them. Similarly flooded roads pose challenge in distinguishing them from non-flooded roads. When a road is flooded it is almost invisible from a top view to determine its existence. Therefore it would be another difficult job for the network models to detect any flooded roads. Most importantly distinguishing between flooded and non-flooded roads and buildings depends on their corresponding contexts.
\fi

\begin{table}[!htp]
	\centering
	\caption{Number of images and instances corresponding to different classes.}\label{table:harvey-data-stat}
	\begin{tabular}{l c c}
	\hline
		Object Class & Images & Instances\\\hline
		\hline
		Building-flooded & 275 & 3573 \\
		Building-non-flooded & 1272 & 5373\\
		Road-flooded & 335 & 649 \\
		Road-non-flooded & 1725 & 3135\\
		Vehicle & 1105 & 6058\\
		Pool & 676 & 1421\\ 
		Tree & 2507 & 25889 \\
		Water & 1262 & 1784\\\hline
	\end{tabular}
\end{table}

%%---------- Argho-------------------------
\subsection{VQA task}
%\input{dataset}

% We collected images from the flood-affected areas after \textit{Hurricane Harvey} with the assistance of an unmanned aerial vehicle ( UAV). Such images are very high in resolution, making them unique compared to other data sets for natural disasters. The post-flooded damages to affected areas are demonstrated in all the images. There are several objects (e.g. building, road) and related attributes ( e.g. state of an object such as flooded or non-flooded after Hurricane Harvey) represented by these images. For the preparation of Semantic Segmentation and Visual Question Answering datasets these attributes are considered.

% \section{Dataset Preparation for Visual question answering}
%Damage assessment after any natural calamities done by querying about the changes  in object's condition. 
To provide VQA framework,  we focus  on generating questions related to 
the building, road, and  entire image as a whole for our \textit{FloodNet} dataset. By asking questions related to these object we can assess the damages and understand the situation very precisely. 
 Attribute associated with aforementioned objects  can be identified from the Table \ref{table:harvey-data-stat}. For the FloodNet-VQA dataset, 
 $\sim{11,000}$ 
question-image pairs  are considered while training VQA networks. All the questions are created manually. Each image has an average of 3.5 questions. Each of the questions is designed to provide answers which are connected to  the local and global regions of images. In Figure \ref{fig:data_rep}, some sample questions-answer pairs are presented from our dataset. 

% \begin{table}[h]
% \centering
% \caption{Object with associated Attributes}
% \begin{tabular}{|c|c|}
% \hline
% \textbf{Object} & \textbf{Associated Attribute}                                  \\ \hline
% Building        & Flooded and Non-Flooded \\ \hline
% Road            & Flooded and Non-Flooded                        \\ \hline
% Entire Image           & Flooded and Non-Flooded  \\ \hline
% \end{tabular}
% \end{table}
 
\subsubsection{Types of Question}

\begin{figure*}
\centering
        \includegraphics[scale=.6]{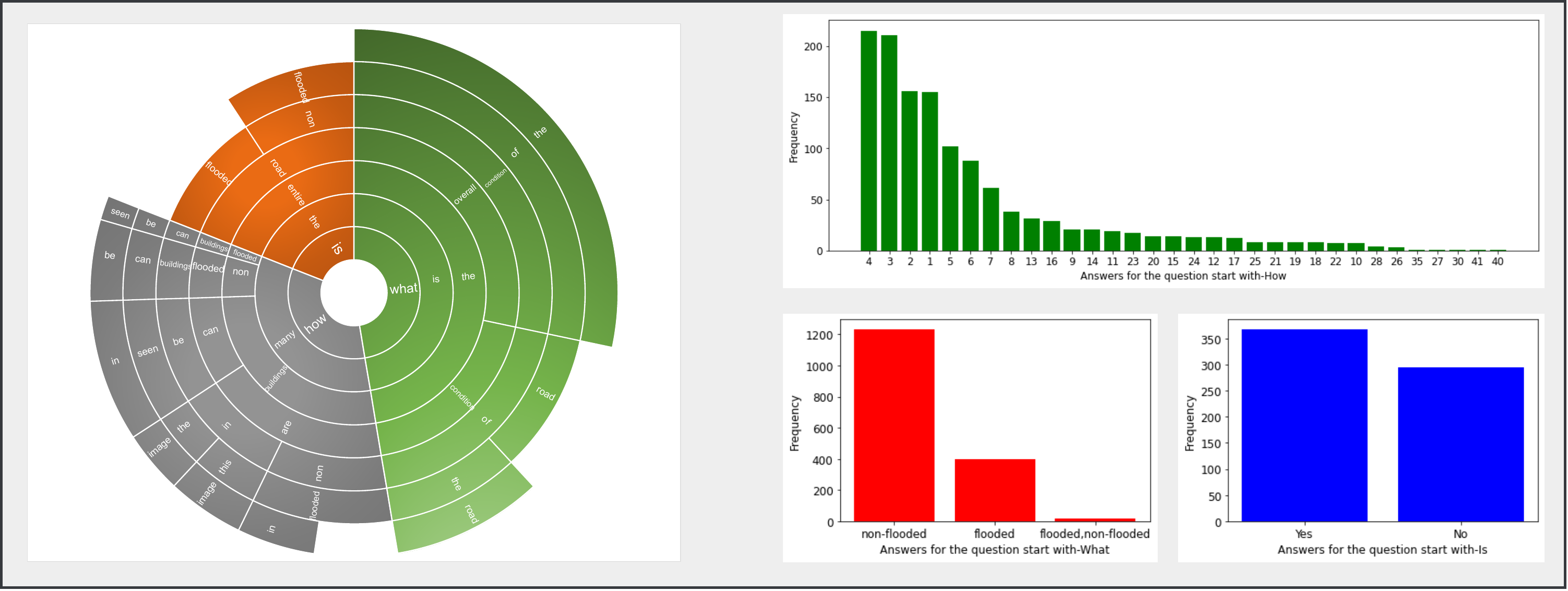}

    \caption{ VQA Data Statistics: Left figure represents the distribution of the question pattern  based on starting word; Right-top, Right-bottom figures describes the distribution of possible answers for different question types }
    \label{Fig:vqa_ques}
\end{figure*}

Questions are divided into a three-way question group, 
namely \textit{``Simple Counting"}, \textit{``Complex Counting"}, and \textit{``Condition Recognition"}. In the  Figure \ref{Fig:vqa_ques}, distribution of the question pattern  based on the first words of the questions is given. All of the questions start with a word belongs to the set \{How, Is, What\}. Maximum length of question is 11.

% inkscape -D .svg  -o image.pdf --export-latex

In the \textit{Simple Counting} problem, we  ask about an object's frequency of presence (mainly building) in an image, regardless of the attribute (e.g. \textit{How many buildings are in the images?}). Both flooded and non-flooded buildings can appear in a picture in several cases (e.g. bottom image from Figure \ref{fig:data_rep}). %However, for a given image, our main focus is to count the total number of buildings (flooded + non-flooded). ``How many ..." begins with the pattern of these questions. 

 \par The question type \textit{Complex Counting} is specifically intended to count the number 
of a particular building attribute (e.g. \textit{How many \textbf{flooded / non-flooded} buildings are in the images?}) 
We're interested in counting only the  flooded or 
non-flooded buildings from this type of query. 
In comparison to simple counting, a high-level understanding of the 
the scene is important for answering this type of question. 
%As all images are taken from a high altitude, this 
%form of query is more difficult to answer even for a person. 
This type of question also starts with the word ``How".
 
 \par \textit{Condition Recognition} questions investigate the condition of the entire image as a whole or the road. This type of question is divided into three sub-categories. One category  deals with the condition of road by asking questions such as   ``What is the condition of the road?". Second one seeks the condition of the entire image  by asking questions like ``What is the overall condition of the entire image?". ``Yes/No" type question is categorised as the third sub-category of the \textit{Condition Recognition}. ``Is the road flooded?", ``Is the road non-flooded" are some of the examples from this sub-category. Starting word for this type of question is either ``Is" or ``What".

\subsubsection{Types of Answer}

%\par Distribution of answers to those three kinds of questions that 
%are generated from the associated attributes of each object are given in Figure \ref{Fig:vqa_ques}. 
% Attributes such as flooded, non-flooded are connected from each image 
% to each object by taking into account the flooded environment. 
% In figure # we have highlighted some scenario that shows the 
% flooded conditions based on which we associated objects with the attributes.

\begin{table}[!htp]
\centering
\caption{Possible Answers for Three Types of Questions}
\vspace{.2cm}
\scalebox{.65}{
\begin{tabular}{c c}
\hline
\textbf{Question Type} & \textbf{Possible Answer}                                  \\ \hline
\hline
Simple Counting        & \{1,2,3,4...\} \\ 
Complex Counting            & \{1,2,3,4...\}                        \\ 
Condition of Road\\(sub-category of Condition Recognition)          & Flooded , Non-Flooded, Flooded \& Non-Flooded  \\ 
Condition of Entire Image\\(sub-category of Condition Recognition)          & Flooded , Non-Flooded  \\ 
Yes/No-Type Question\\(sub-category of Condition Recognition)          & Yes, No  \\ \hline

\end{tabular}}
\label{tab:answer}
\end{table}

%Buildings are annotated as non-flooded if floodwater is not present near any building. 
%On the contrary, if there exists floodwater near any building, we 
%have marked those buildings as flooded. 

Both flooded and non-flooded buildings 
can exist in any image. For complex counting problem, we only count either the flooded or non-flooded buildings from a given image-question pair.   Roads are also annotated as flooded or non-flooded.  Second image from the Figure \ref{fig:data_rep} depicts both flooded and non-flooded roads. Thus, the answer for the question like ``What is condition of road?" for this kind of images will be both `flooded and non-flooded'. Furthermore, entire image may be graded as flooded or non-flooded. Table \ref{tab:answer} refers to the possible answers for three types questions and     from Figure \ref{Fig:vqa_ques}, we can see the possible answer distribution for different types of question. Most frequent  answers for counting problem, in general, are  `4, 3, 2, 1' whereas `27, 30, 41, 40' are the less frequent answers. For \textit{Condition Recognition} problem, `non-flooded, yes' are the most common answers.

\begin{figure*}[t]
	\begin{center}
		\includegraphics[width=\linewidth]{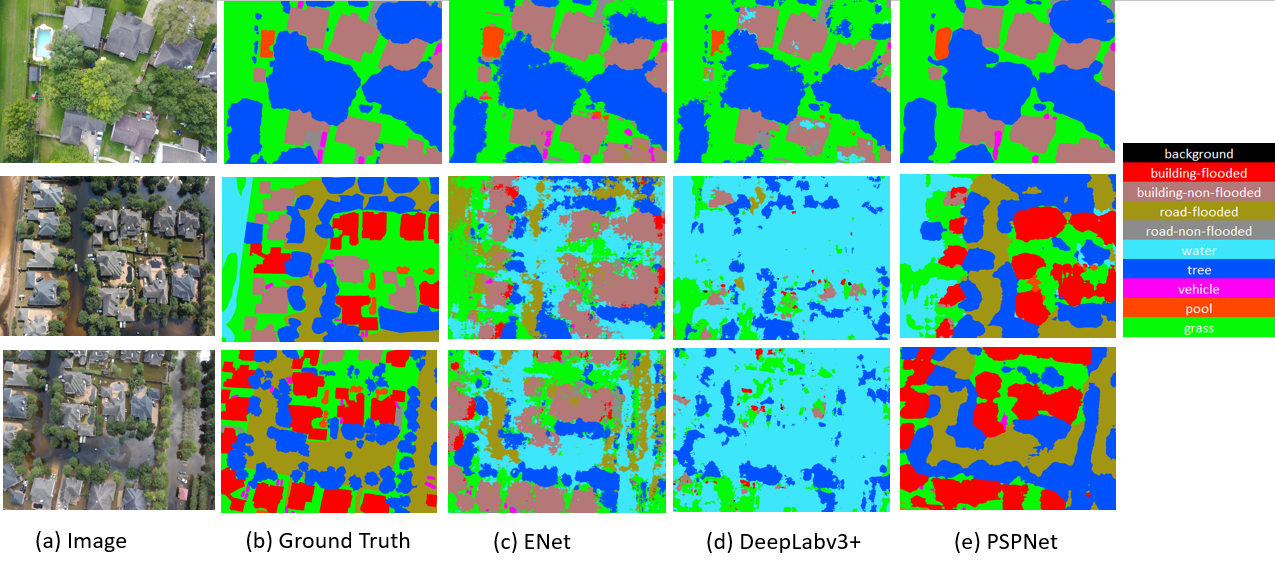}
	\end{center}
	\caption{Visual comparison on FloodNet test set for Semantic Segmentation.}
	\label{fig:vis-compare-segmentation-models-all-cls}
\end{figure*}

%% results on segmentation experiment (All-cls)
\begin{table*}[!htp]
	\centering
	\caption{Per-class results on FloodNet testing set.}\label{table:harvey-segmentation-test-perfor-table}
	\begin{adjustbox}{width=\textwidth}
	\begin{tabular}{l c c c c c c c c c| c}
	\hline
		Method & Building Flooded & \begin{tabular}{@{}c@{}}Building Non \\  Flooded\end{tabular} & Road Flooded &  \begin{tabular}{@{}c@{}}Road Non \\  Flooded \end{tabular} & Water & Tree & Vehicle & Pool & Grass & mIoU\\\hline
		\hline
		ENet\cite{paszke2016enet} & 6.94 & 47.35 & 12.49 & 48.43 & 48.95 & 68.36 & 32.26 & 42.49 & 76.23 & 42.61\\
		DeepLabV3+\cite{chen2018encoder} & 32.7 & 72.8 & 52.00 & 70.2 & 75.2 & 77.00 & 42.5 & 47.1 & 84.3 & 61.53\\ 
		PSPNet\cite{zhao2017pyramid} & \bftab 68.93 & \bftab 89.75 & \bftab 82.16 & \bftab 91.18 & \bftab 92.00 & \bftab 89.55 & \bftab 46.15 & \bftab 64.19 & \bftab 93.29 & \bftab 79.69\\ \hline
	\end{tabular}
	\end{adjustbox}
\end{table*}

\begin{table*}[t]

\centering
\caption{Accuracy table for Baseline VQA Algorithms}
\vspace{.2cm}
%\scalebox{.7}{
\begin{adjustbox}{width=\textwidth}
\begin{tabular}{ccc|c|c|c|c|c|}
\cline{4-8}
                                                                               &                                         &                           & \multicolumn{2}{c|}{Counting Problem}                                                                                                                                   & \multicolumn{3}{c|}{Condition Recognition}                                                                                                                                                                                                                    \\ \hline
\multicolumn{1}{|c|}{\textbf{Method}}                                          & \multicolumn{1}{c|}{\textbf{Data Type}} & \textbf{Overall Accuracy} & \textbf{\begin{tabular}[c]{@{}c@{}}Accuracy for \\ 'Simple Counting'\end{tabular}} & \textbf{\begin{tabular}[c]{@{}c@{}}Accuracy for\\ 'Complex Counting'\end{tabular}} & \textbf{\begin{tabular}[c]{@{}c@{}}Accuracy for \\ 'Yes/No'\end{tabular}} & \textbf{\begin{tabular}[c]{@{}c@{}}Accuracy for \\ "Entire Image\\ Condition"\end{tabular}} & \textbf{\begin{tabular}[c]{@{}c@{}}Accuracy for \\ "Road\\ Condition"\end{tabular}} \\ \hline
\multicolumn{1}{|c|}{\multirow{2}{*}{Concatenation of Features \cite{zhou2015simple}}}                & \multicolumn{1}{c|}{Validation}         & 0.41                      & 0.04                                                                               & 0.03                                                                               & 0.017                                                                     & 0.86                                                                                        & 0.9                                                                                 \\ \cline{2-8} 
\multicolumn{1}{|c|}{}                                                         & \multicolumn{1}{c|}{Testing}            & 0.42                      & 0.04                                                                               & 0.03                                                                               & 0.17                                                                      & 0.86                                                                                        & 0.9                                                                                 \\ \hline
\multicolumn{1}{|c|}{\multirow{2}{*}{Element-wise Multiplication of Features \cite{antol2015vqa}}} & \multicolumn{1}{c|}{Validation}         & 0.69                      & 0.28                                                                               & 0.27                                                                               & 0.86                                                                      & 0.96                                                                                        & 0.97                                                                                \\ \cline{2-8} 
\multicolumn{1}{|c|}{}                                                         & \multicolumn{1}{c|}{Testing}            & 0.68                      & 0.25                                                                               & 0.21                                                                               & 0.84                                                                      & 0.96                                                                                        & 0.97                                                                                \\ \hline
\multicolumn{1}{|c|}{\multirow{2}{*}{SAN \cite{yang2016stacked}}}                                     & \multicolumn{1}{c|}{Validation}         & 0.63                      & 0.34                                                                               & 0.28                                                                               & 0.51                                                                      & 0.95                                                                                        & 0.97                                                                                \\ \cline{2-8} 
\multicolumn{1}{|c|}{}                                                         & \multicolumn{1}{c|}{Testing}            & 0.63                      & 0.26                                                                               & 0.24                                                                               & 0.54                                                                      & 0.94                                                                                        & 0.97                                                                                \\ \hline
\multicolumn{1}{|c|}{\multirow{2}{*}{MFB with Co-Attention \cite{yu2017multi}}}                   & \multicolumn{1}{c|}{Validation}         & 0.72                      & 0.31                                                                               & 0.28                                                                               & 0.98                                                                      & 0.96                                                                                        & 0.97                                                                                \\ \cline{2-8} 
\multicolumn{1}{|c|}{}                                                         & \multicolumn{1}{c|}{Testing}            & \textbf{0.73}             & \textbf{0.29}                                                                      & \textbf{0.26}                                                                      & \textbf{0.99}                                                             & \textbf{0.97}                                                                               & \textbf{0.99}                                                                       \\ \hline
\end{tabular}
\end{adjustbox}
%}
\label{tab:result_vqa}

\end{table*}

\section{Experiments}
\label{sec:exp}
\par To understand the usability of these images for flood detection, we majorly carry out three tasks, which are Image Classification, Semantic Segmentation, and Visual Question Answering (VQA). We begin with classifying the FloodNet data into Flooded and Non-Flooded images, then we detect specific regions of flooded buildings, flooded roads, vehicles etc. through semantic segmentation networks. Finally, we carry out VQA on this dataset. For all of our tasks, we use NVIDIA GeForce RTX 2080 Ti GPU with an Intel Core i9 processor. 

\par For image classification, we used three state-of-the-art networks i.e. InceptionNetv3 \cite{szegedy2016rethinking}, ResNet50 \cite{he2016deep}, and Xception \cite{chollet2017xception} as base models to classify the images into Flooded and Non-Flooded categories. These networks have significantly contributed to the field of Computer Vision by introducing a unique design element, such as the residual blocks in ResNet, the multi-scale architecture in InceptionNet and depthwise separable convolutions in Xception. For our classification task, the output from these base models was followed by a Global Average Pooling Layer, a fully connected layer with 1024 neurons having Relu Activation, and finally by two neurons with Softmax activation. We initialized our networks with ImageNet \cite{deng2009imagenet} weights and trained them for 30 epochs, with 20 steps for every epoch, using binary cross entropy loss. %The dataset was divided into 876 images for training and 111 images for testing, with all images being resized to 224 $\times$ 224.

\par For semantic segmentation, we implemented three methods, i.e. PSPNet \cite{zhao2017pyramid}, ENet \cite{paszke2016enet}, and DeepLabv3+ \cite{chen2018encoder}; and evaluate their performance on FloodNet dataset. For implementing PSPNet, ResNet101 was used as backbone. We used “poly” learning rate with base learning rate 0.0001. Momentum, weight decay, power, and weight of the auxiliary loss were set to 0.9, 0.0001, 0.9, and 0.4 respectively. For ENet we used 0.0005 and 0.1 for learning rate and learning rate decay respectively. Weight decay was set to 0.0002. Similarly for DeepLabv3+ we used poly learning rate with base learning rate 0.01. We set weight decay to 0.0001 and momentum to 0.9. For image augmentation we used random shuffling, scaling, flipping, and random rotation which helped the models avoid overfitting. From different experiments it was proved that larger “crop size” and “batch size” improve the performance of the models. During training, we resized the images to 713 $\times$ 713 since large crop size is useful for the high resolution images. For semantic segmentation evaluation metric we used mean IoU (mIoU).

For Visual Question Answering,  simple baselines (concatenation/element-wise product of image and text features) and Multimodal Factorized Bilinear (MFB) with co-attention \cite{yu2017multi}, Stacked Attention Network \cite{yang2016stacked} have been considered for this study. All of these models are configured according to our dataset. For image and question feature extraction, respectively, VGGNet (VGG 16) and Two-Layer LSTM are taken into account.  Feature vector from last pooling layer of the VGGNet  and 1024-D  vector from the last word of Two-Layer LSTM are considered as the image and question vectors respectively. Dataset is splited into training, validation and testing data. All the images are resized to 224 $\times $ 224 and questions are tokenized. By considering  cross-entropy  loss, all the models are optimized by stochastic gradient descent ( SGD) with batch size 16. In the training phase, models are validated by validation dataset via early stopping criterion with patience 30.
% \input{table/result}
%%--------------------Debvrat-----------------------------

\subsection{Image Classification Analysis}

The classification accuracies of the three networks are shown in Table \ref{tab:classification_results}. From this table, it can be seen that the highest performance on the test set was given by ResNet. The residual architecture of ResNet has successfully helped in classifying the test images into Flooded and Non-Flooded, as compared to the other networks. Even though Xception and InceptionNet have a much wider architecture and show higher classification accuracy on ImageNet data, this is not the case for FloodNet dataset. 

Therefore, networks which give high accuracy on everyday images such as those of ImageNet can not really be used to detect image features from aerial datasets which contain more complex urban and natural scenes. Thus, there is a need to design separate novel architectures which can effectively detect urban disasters.

\begin{table}[h!]
    \centering
     \caption{Classification accuracy of three state-of-the-art networks on FloodNet dataset}
    \begin{tabular}{c|c|c}
         Model & Training Accuracy & Test Accuracy  \\
         \hline
         InceptionNetv3\cite{szegedy2016rethinking} & 99.03 \% & 84.38\% \\
         ResNet50\cite{he2016deep} & 97.37\%  & 93.69\% \\
        Xception\cite{chollet2017xception} & 99.84 \% & 90.62\%
    \end{tabular}
   
    \label{tab:classification_results}
\end{table}

%%--------------------Tashnim-----------------------------
% \subsection{Semantic Segmentation Experiment}
% We implement three segmentation methods, PSPNet \cite{zhao2017pyramid}, ENet \cite{paszke2016enet}, and DeepLabv3+ \cite{chen2018encoder}, and evaluate their performance on FloodNet dataset. For implementing PSPNet, resnet101 has been used as backbone. We use “poly” learning rate with base learning rate 0.0001. Momentum, weight decay, power, and weight of the auxiliary loss are set to 0.9, 0.0001, 0.9, and 0.4 respectively. For ENet we use 0.0005 and 0.1 for learning rate and learning rate decay respectively. Weight decay is set to 0.0002. Similarly for DeepLabv3+ we use poly learning rate with base learning rate 0.01. We set weight decay to 0.0001 and momentum to 0.9. For augmentation we use random shuffling, scaling, flipping, and random rotation which help models to avoid overfitting. From different experiments it is proved that larger “crop size” and “batch size” improve the performance of the models. During training, we resize the images to 713 $\times$ 713 since large crop size is useful for the high resolution images. For semantic segmentation evaluation metric we use mean IoU (mIoU).

%%-----------------------Tashnim-----------------------------------------
\subsection{Semantic Segmentation Performance Analysis}
Semantic segmentation results of ENet, DeepLabv3+, and PSPNet are presented in Table \ref{table:harvey-segmentation-test-perfor-table}. From the segmentation experiment it is evident that detecting small objects like vehicles and pools are the most difficult tasks for the segmentation networks. Then flooded buildings and roads are the next challenging tasks for all three models. Among all of the segmentation models, PSPNet performs best in all classes. It is interesting to note that although DeepLabv3+ and PSPNet collect global contextual information, still their performance on detecting flooded building and flooded roads are still low, since distinguishing between flooded and non-flooded objects heavily depend on respective contexts of the classes.

%-------------------------------------------------------------------------

\subsection{Visual Question Answering Performance Analysis}

From the Table \ref{tab:result_vqa}, we can identify that counting problem (simple and complex) is very challenging compare to task of condition recognition. Many objects are   very small  which makes it very difficult even for human to count. Accuracy for \textit{`Condition Recognition'} category is high. This is because it is not difficult to recognize the condition of whole images as well as roads as they are pictured in a larger ratio given the overall size of an image. MFB with co-attention \cite{yu2017multi} outperforms all the other methods for all types of question.     

%%----------------------Tashnim------------------
\section{Discussion and Conclusion}
\label{sec:final}
In this paper, we introduce the FloodNet dataset for post natural disaster damage assessment. We describe the dataset collection procedure along with different features and statistics. The UAV images provide high resolution and low altitude dataset specially significant for performing computer vision tasks. The dataset is annotated for classification, semantic segmentation, and VQA. We perform three computer vision tasks including image classification, semantic segmentation, and visual question answering and in-depth analysis have been provided for all three tasks.

Although UAVs are cost effective and prompt solution during any post natural disaster damage assessment, several challenges have been posed by FloodNet dataset collected using UAVs. Among all the existing classes, vehicles and pools are the smallest in shape and therefore would be difficult for any network models to detect them. Segmentation results from Table \ref{table:harvey-segmentation-test-perfor-table} supports the task difficulty in identifying small objects like vehicles and pools. Besides detecting flooded building is another prime challenge. Since UAV images only include top view of a building, it is very difficult to estimate how much damages are done on that building. \iffalse Horizontal view brings most information about a building's current condition after the flood, but the UAV images provide only top views. We are defining a building to be flooded if at least one end or any outside wall of the building is touching the flood water. From an UAV image it is very difficult to verify whether flood water is touching at least one end of a building which makes distinguishing between flooded and non-flooded buildings a challenging task for any network model. \fi Segmentation models do not perform well in detecting flooded buildings. Similarly flooded roads pose challenge in distinguishing them from non-flooded roads and results from segmentation models prove that. Most importantly distinguishing between flooded and non-flooded roads and buildings depends on their corresponding contexts and current state-of-art models are still lacking good performance in computer vision tasks performed on FloodNet.
To the best of our knowledge this is the first time where these three crucial computer vision tasks have been addressed in a post natural disaster dataset together. The experiments of the dataset show great challenges and we strongly hope that FloodNet will motivate and support the development of more sophisticated models for deeper semantic understanding and post disaster damage assessment.

\section{Acknowledgment}

This work is partially supported by Microsoft and Amazon.

{\small
\bibliographystyle{ieee_fullname}
\bibliography{cvpr-harvey}
}

\end{document}